\pdfoutput=1

\documentclass[11pt]{article}

\usepackage[]{acl}

\usepackage{times}
\usepackage{latexsym}
\usepackage{multirow}
\usepackage{graphicx}
\usepackage{changepage}
\usepackage[T1]{fontenc}

\usepackage[utf8]{inputenc}
\usepackage[normalem]{ulem}

\usepackage{microtype}

\usepackage{booktabs}
\usepackage{url}
\usepackage{color}

\title{IndicBART: A Pre-trained Model for Indic Natural Language Generation}

\author{Raj Dabre$^1$ \quad
        Himani Shrotriya$^2$ \quad
        Anoop Kunchukuttan$^3$ \\
        \textbf{Ratish Puduppully}$^4$ \quad \textbf{Mitesh M. Khapra}$^5$ \quad \textbf{Pratyush Kumar}$^6$ \\
        National Institute of Information and Communications Technology$^{1}$ \quad IIT Madras$^{2,5,6}$ \\
        Microsoft$^{3,6}$ \quad
        University of Edinburgh$^{4}$ \\
        {\tt $^1$raj.dabre@nict.go.jp} \quad
        {\tt $^2$cs20m024@smail.iitm.ac.in} \\
        {\tt $^3$ankunchu@microsoft.com
} \quad {\tt $^4$r.puduppully@sms.ed.ac.uk
} \\ {\tt $^5$miteshk@cse.iitm.ac.in} \quad {\tt $^6$pratykumar@microsoft.com}
}

\begin{document}
\maketitle
\begin{abstract}
In this paper, we study pre-trained sequence-to-sequence models for a group of related languages, with a focus on Indic languages. We present IndicBART, a multilingual,  sequence-to-sequence pre-trained model focusing on 11 Indic languages and English. IndicBART utilizes the orthographic similarity between Indic scripts to improve transfer learning between similar Indic languages. We evaluate IndicBART on two NLG tasks: Neural Machine Translation (NMT) and extreme summarization. Our experiments on NMT and extreme summarization show that a model specific to related languages like IndicBART is competitive with large pre-trained models like mBART50 despite being significantly smaller. It also performs well on very low-resource translation scenarios where languages are not included in pre-training or fine-tuning. Script sharing, multilingual training, and better utilization of limited model capacity contribute to the good performance of the compact IndicBART model.



\end{abstract}

\section{Introduction}

Recently, there has been  significant progress in deep learning based natural language generation (NLG) for machine translation, abstractive summarization, 
data-to-text generation,
etc. due to the adoption of attention-based sequence-to-sequence (S2S) models (conditional language models) \cite{DBLP:journals/corr/WuSCLNMKCGMKSJL16,paulus2018a,DBLP:conf/aaai/Puduppully0L19}. Pre-trained S2S models have been shown to be useful to improve performance on various NLG tasks \cite{rothe-etal-2020-leveraging,kale-rastogi-2020-text,lewis-etal-2020-bart}. Specifically, multilingual pre-trained S2S models jointly trained on monolingual corpora from multiple languages such as mBART25 \cite{liu-etal-2020-multilingual-denoising}, mBART50 \cite{tang2020multilingual} and mT5 \cite{xue-etal-2021-mt5} have seen increased adoption and low-resource languages have benefitted from cross-lingual transfer. However, these massively multilingual massive (M3) models have major limitations. They serve only a few of the world's languages (<100 languages), the pre-training corpora are dominated by high-resource languages, the vocabulary representation for low-resource languages is inadequate, and the models are large, making them expensive and slow to train, fine-tune and decode.  

An alternative approach is to build pre-trained S2S models for a group of related languages. Previous work has shown the benefits of pre-trained language models as well as NMT models that cater to a set of related languages \cite{kakwani-etal-2020-indicnlpsuite,tan18,khanuja2021muril,reid-etal-2021-afromt}. Owing to their public availability, these models have seen heavy adoption\footnote{Over 10,000 downloads for MuRIL (\url{https://huggingface.co/google/muril-base-cased}) and IndicBERT (\url{https://huggingface.co/ai4bharat/indic-bert}).}. However, such a study on multilingual pre-trained S2S models for Indic languages is missing in the literature. In this work, we address this gap in the literature by studying multilingual pre-trained S2S models for Indic languages. 


The result of this study is \textit{IndicBART}, a multilingual pre-trained sequence to sequence model specifically trained for Indic languages, which are spoken by more than a billion users\footnote{\url{https://en.wikipedia.org/wiki/Demographics_of_India}}. It \textbf{supports English and 11 Indian languages} including 7 Indo-Aryan (Assamese, Bengali, Gujarati, Hindi, Marathi, Oriya, Punjabi) and 4 Dravidian (Kannada, Malayalam, Tamil, Telugu) languages. Of these, mBART25, mBART50 and mT5 support only 2, 7 and 9 languages respectively. There are linguistic similarities between the two language families on account of contact relatedness resulting from geographical colocation. Within, the two language families there are genetic relations between languages due to them being derived from common ancestor languages\footnote{\url{https://en.wikipedia.org/wiki/Proto-Indo-Aryan_language}}\footnote{\url{https://en.wikipedia.org/wiki/Proto-Dravidian_language}}. Due to this, the Indian subcontinent is considered to be a \textit{linguistic area} or \textit{sprachbund} \cite{emeneau1956india}. There is evidence that such contact-relatedness can result in positive cross-lingual transfer for NLP applications like NMT \cite{goyal-etal-2020-contact}. Hence, we train a single model for all Indic languages.  It is a \textbf{compact model with just 244M parameters}, which is much smaller than the M3 models such as mBART50 and mT5(-base) which contain 611M and 580M parameters respectively. We also propose a variant of IndicBART, i.e. \textbf{IndicALBART}, that is highly compact with just \textbf{97M parameters}. 

We compare IndicBART with M3 models on two downstream generation tasks: machine translation and extreme summarization \cite{narayan-etal-2018-dont}. The results indicate that IndicBART is competitive or better by up to 2 BLEU/ROUGE compared to M3 models like mBART50. IndicBART also performs well in the following zero-shot scenarios: (a)~on languages not included in pre-training, and  (b)~languages for which there is no fine-tuning data. 


The following aspects of the IndicBART model contribute to its strong performance and increased language coverage within the Indic group vis-à-vis M3 models, while being highly compact:

\noindent \textbf{1.} It is trained on a smaller set of related languages, which reduces model capacity requirements. Moreover, available model capacity is effectively utilized, since transfer learning works when languages share linguistic features and data represents shared topical themes.

\noindent \textbf{2.} It is trained on the largest publicly available Indic language corpora, IndicCorp  \cite{kakwani-etal-2020-indicnlpsuite}, which includes large, high-quality news crawls for Indian languages as well as English content from Indian websites - thus being representative of Indian English and topics. 

\noindent \textbf{3.} We utilize the orthographic similarity between Indic scripts \cite{kunchukuttan-etal-2018-leveraging} to map all the Indic language data to a single script, effectively reducing the number of scripts from 9 to 1 (each script having approximately 50 characters). This increases the shared subwords in the vocabulary, and we observe that single script models enable better cross-lingual transfer while fine-tuning. Since subwords embeddings consume a significant fraction of the parameter space, single script models also better utilize available vocabulary budget\footnote{Where mBART-25 and mBART-50 have vocabularies of 250K subwords to accommodate 25 to 50 languages, IndicBART has a vocabulary of 64K subwords which is 4 times smaller.}.

\noindent \textbf{4.} Extremely compressed pre-trained S2S models (IndicALBART) suitable for deployment can be trained by sharing parameters across layers of the transformer layers. For related languages, we show  compressed pre-trained models are competitive with full models on downstream tasks when fine-tuned on distilled data. 

The IndicBART model and its variants, along with details on how to fine-tune them, can be accessed at \url{https://github.com/AI4Bharat/indic-bart/}. We also release the models on the HuggingFace model hub at \url{https://huggingface.co/ai4bharat/IndicBART} and \url{https://huggingface.co/ai4bharat/IndicBARTSS}. Models are available under an MIT license to spur further innovation in NLG for Indic languages and study of pre-trained S2S models for related languages. 

\section{Related Work}

\noindent \textbf{Pre-trained models.} Pre-trained models learned using self-supervised objectives and large monolingual corpora have contributed to rapid advances in NLU  \cite{devlin-etal-2019-bert} and NLG \cite{lewis-etal-2020-bart}. Following initial work on English pre-trained models, multilingual pre-trained models have been proposed for NLU \cite{devlin-etal-2019-bert,conneau-etal-2020-unsupervised}  as well as NLG \cite{liu-etal-2020-multilingual-denoising,tang2020multilingual,xue-etal-2021-mt5} supporting around 100 languages. These pre-trained M3 models have proven to be very useful in improving NLG performance in low-resource settings, especially for applications other than translation.

\noindent \textbf{Language group-specific models.} The proposed IndicBART model is also a multilingual pre-trained S2S model, similar in architecture and training to mBART. However, in contrast to mBART and mT5, the proposed IndicBART caters specifically to Indic languages. While language-group specific NLU language models like IndicBERT \cite{kakwani-etal-2020-indicnlpsuite}  and MuRIL \cite{khanuja2021muril} and NMT models \cite{tan18} have been proposed, ours is one of the first efforts to create a pre-trained S2S model for a specific language group (and the first for Indic languages). AfroMT \cite{reid-etal-2021-afromt} is a concurrent effort focussed on African languages and low monolingual corpora scenarios belonging to various language families. However, AfroMT heavily relies on synthetic data, which may not reflect the true data distribution across languages. Furthermore, AfroMT effort is focussed only on MT, whereas we investigate IndicBART on an additional NLG task - abstractive summarization. Interestingly, the publicly available group-specific language models (IndicBERT and MuRIL) both cater to Indic languages, pointing to perceived need for Indic language specific models. 

\noindent \textbf{Language relatedness.}
Language-group specific models are motivated from previous work that emphasizes the role of language relatedness in cross-lingual transfer for NMT \cite{nguyen-chiang-2017-transfer,dabre-etal-2017-empirical,aharoni-etal-2019-massively,kudugunta-etal-2019-investigating,dabre2020} and NLU \cite{kakwani-etal-2020-indicnlpsuite,khemchandani-etal-2021-exploiting,dhamecha-etal-2021-role}. We use a single script for representing Indic data since orthographic similarity between Indic languages has been utilized to represent data in a common script and improve cross-lingual transfer for machine transliteration \cite{kunchukuttan-etal-2018-leveraging}, machine translation \cite{dabre2018nict,goyal2020contact,DBLP:journals/corr/abs-2104-05596} and NLU \cite{khemchandani-etal-2021-exploiting,dhamecha-etal-2021-role}. 

\noindent \textbf{Parameter Sharing and Distillation.} Parameter sharing across layers has shown promise for NMT \cite{Dabre_Fujita_2019} and pre-trained LMs \cite{conf/iclr/LanCGGSS20} in building compressed models while maintaining end-task performance. The IndicALBART model proposed in this work is the first model to explore parameter-sharing across layers for pre-trained S2S models. For NMT models trained from scratch, sequence-to-sequence distillation \cite{kim-rush-2016-sequence} has been shown as an effective way to transfer knowledge to smaller models, while training large models on distilled data (a form of self-training) has been shown to improve translation quality \cite{dabre-fujita-2020-combining}. Our results indicate that these results hold when fine-tuning on pre-trained S2S models as well.

\section{IndicBART}
The \textbf{IndicBART} model is conceptually based on the mBART25/50 model family, which are Transformer models \cite{vaswani2017attention} trained on monolingual corpora with masked span reconstruction objective. We refer the readers to the mBART literature \cite{lewis-etal-2020-bart,liu-etal-2020-multilingual-denoising} for architectural details and highlight specific details and differences from the mBART25/50 setup. 


\subsection{Design Considerations for IndicBART}
Considerations that drove our model choices are:

\noindent \textbf{Compactness}: The model should be compact given our focus on a smaller set of related languages, as well as to accelerate training and fine-tuning. Such a model will be usable by a larger base of users with limited computational resources. 

\noindent \textbf{Content Relevance}: In addition to Indian languages, we include English since transfer-learning from English is a natural use case, and English is widely used in the Indian subcontinent. We also use English content from the Indian subcontinent to reflect relevant content. 

\noindent \textbf{Leveraging Relatedness}: We utilize orthographic similarity between Indian languages, most of which use abugida scripts derived from the Brahmi script. The logical character set has high overlaps, though each script has its own code-point range in the Unicode standard \cite{kunchukuttan-etal-2018-leveraging}. We map all the data to Devanagari, enabling better transfer learning\footnote{There is a substantial amount of shared vocabulary between Indian languages written in different scripts. Mapping scripts to Devanagari enables direct sharing of vocabulary, leading to improved transfer learning.} with a more compact vocabulary compared to mBART.

\subsection{Model and Training Details} 
IndicBART uses (N=) 6 encoder and decoder layers with  hidden and filter sizes of 1024 and 4096, respectively, and 16 attention heads (244M parameters). Similar to mBART, we mask (p=)35\% of the words in each sentence by randomly sampling a span length according to a
Poisson distribution ($\lambda=3.5$).  We use dropouts of 0.1, label smoothing of 0.1, Adam optimizer with a maximum learning rate of 0.001, weight decay of 0.00001, linear learning rate warm-up and decay with 16,000 warm-up steps, batch sizes of 4096 tokens. We train for 750,000 iterations on 48 NVIDIA V-100 GPUs, corresponding to roughly 2 epochs, taking around 5 days\footnote{Longer training was limited by the availability of many GPUs simultaneously.}. In comparison, mBART25/50 models need much longer time (2+ weeks) on 256 GPUs.

To explore more compressed pre-trained models, we train \textbf{IndicALBART}, a variant of IndicBART with cross-layer parameter sharing, i.e., sharing parameters across layers. For ablation studies on the impact of single script representation, we also train a variant of IndicBART with a 64K vocabulary using the original scripts, which we call separate script IndicBART (SSIndicBART). 

 The models have been trained with the YANMTT toolkit\footnote{https://github.com/prajdabre/yanmtt} \cite{dabre2021yanmtt} which is based on the mBART implementation of the HuggingFace Transformers library \cite{wolf-etal-2020-transformers}.

\subsection{Training Data and Pre-processing} 
We train the IndicBART model on the IndicCorp (IC) dataset \cite{kakwani-etal-2020-indicnlpsuite} which contains 11 Indic languages and English. The Indic languages are: Assamese (as), Bengali (bn), Gujarati (gu), Hindi (hi), Kannada (kn), Malayalam (ml), Marathi (mr), Oriya (or), Punjabi (pa), Tamil (ta) and Telugu (te). The corpora statistics are mentioned in Table~\ref{tab:databi} of the appendix. We train the model on a total of approx. 450 million sentences and 9 billion tokens, where corpora sizes are balanced with temperature (T=5) based sampling \cite{wild}. All the Indic language data is represented in a single script, i.e., the Devanagari script using the IndicNLP library\footnote{https://github.com/anoopkunchukuttan/indic\_nlp\_library} \cite{kunchukuttan2020indicnlp}.  We use a vocabulary of 64K subwords learned using SentencePiece \cite{kudo-2018-subword,kudo-richardson-2018-sentencepiece} on randomly sampled 1M raw sentences from the IndicCorp for each language, for a total of 12M sentences. The model is trained at the sentence-level, unlike the mBART50 model, which is trained on contiguous text chunks potentially spanning multiple sentences.

\section{Experiments: NMT} 
Machine Translation is a standard, popular, cross-lingual generation task for which various pre-trained models are evaluated. We compare IndicBART and its variants with mBART50, which should be the most directly comparable model. We study their performance in: (a) low-resource, (b) multilingual and (c) zero-shot training settings. 

\subsection{Models Compared}

We study IndicBART via the following models: 

\noindent\textbf{Models trained from scratch:} We train bilingual (Bi) as well as multilingual many-to-one (M2O) and one-to-many (O2M) transformer models.

\noindent\textbf{Fine-tuned models:} We fine-tune mBART50 (MB50), IndicBART (IB) and its variants namely IndicALBART (IALB) and separate script IndicBART (SSIB). The type of fine-tuning is indicated by +type, which can be Bi, O2M or M2O. If needed, the corpus is indicated by +corpus.

\noindent\textbf{Distilled models:} We use the multilingually fine-tuned IndicBART model and translate the training data source sentences,  which yields distillation data \cite{kim-rush-2016-sequence}. We use this data to train M2O and O2M models from scratch, as well as by fine-tuning on mBART50, IndicBART and IndicALBART. This was motivated by \citet{dabre-fujita-2020-combining} who show that the distillation data generated using models employing transfer learning significantly improves the performance of compact models for low-resource languages.

\subsection{Datasets and Preprocessing}
The statistics of training corpora are in Table~\ref{tab:databi} in the appendix.

\noindent \textbf{Training:}
For a low-resource setting (LR), we use the PMI subset \cite{haddow2020pmindia} of the WAT 2021 MultiIndicMT\footnote{http://lotus.kuee.kyoto-u.ac.jp/WAT/indic-multilingual} \cite{nakazawa-etal-2021-overview} training set for finetuning. This represents an extremely low-resource parallel corpus setting where we expect IndicBART to be the most helpful. We experiment with extending the PMI data (approximately 326K pairs) with the CVIT-PIB (henceforth PIB: 930K pairs) data \cite{siripragrada-etal-2020-multilingual} which is similar in domain to the former. We also use the high-resource, general domain Samanantar corpus \cite{DBLP:journals/corr/abs-2104-05596} (46.2M pairs) to compare with the generalization capabilities of pre-trained models which are fine-tuned with small corpora (PMI, PIB).

\noindent \textbf{Testing:} We use the WAT 2021 MultiIndicMT testset and the FLORES101 devtest \cite{DBLP:journals/corr/abs-2106-03193} for evaluation of our models. Both these test sets are $n$-way parallel (2,390 and 1,012 sentences respectively). The WAT 2021 test set shares the same domain as the training set. The FLORES devtest comes from a different, general domain. We rely on the FLORES dataset to evaluate performance of models trained on the PMI/PIB domain on a more general domain. 

\noindent \textbf{Validation:} We use the WAT2021 development set of 1,000 sentences. 

\noindent \textbf{Preprocessing:} For IndicBART and IndicALBART, we use the Indic NLP library to convert the Indic side of the parallel data to the Devanagari script. For mBART50, only Kannada, Punjabi and Oriya scripts are converted to Devanagari as mBART50 does not support these languages. Results for these are italicized. For separate script IndicBART we do not do script conversion. 

With this setup, we study the benefits of pre-training in low-resource settings (fine-tuned on PMI and PIB) and compare it with high-resource settings (trained on Samanantar) on in-domain (WAT2021) and general (FLORES) test sets. Unless explicitly mentioned, our models are assumed to be trained/fine-tuned/distilled with the PMI training data.

\begin{table*}[t]
\centering
\begin{tabular}{l|c|cccccccccc}
\multirow{1}{*}{\textbf{Model}}     & \multirow{1}{*}{\textbf{\#Params}} &  \textbf{bn} & \textbf{gu} & \textbf{hi} & \textbf{kn}   & \textbf{ml} & \textbf{mr} & \textbf{or}   & \textbf{pa}   & \textbf{ta} & \textbf{te}                                                                                                  \\\cline{1-12}
                         
                         \multicolumn{12}{c}{\textbf{XX-En}}  \\\hline
\multicolumn{12}{c}{\textbf{Bilingual Models}}  \\\hline
\textbf{Bi}        & 78M        & 13.5        & 27.4        & 30.9        & 22.5          & 16.5        & 18.4        & 18.4          & 27.1          & 17.1        & 16.5        \\
\textbf{MB50+Bi}   &  611M   & 23.2        & 35.4        & \textbf{38.3}        & \textit{26.8} & \textbf{29.2}        & \textbf{27.7}        & \textit{27.8} & \textit{35.8} & \textbf{27.1}        & \textbf{30.8}        \\
\textbf{IB+Bi}  & 244M  & \textbf{23.6}        & \textbf{35.5}        & 36.8        & \textbf{31.6}        & 27.9        & 26.8        & \textbf{28.3}        & \textbf{36.3}        & 27.0        & 29.9        \\\hline
\multicolumn{12}{c}{\textbf{Multilingual Models}}  \\\hline
\textbf{M2O}        & 78M       & 18.9        & 24.8        & 27.8        & 23.8          & 21.6        & 20.7        & 21.2          & 26.4          & 20.6        & 21.8        \\

\textbf{MB50+M2O}   &  611M   & \textbf{24.8}        & \textbf{33.9}        & 36.8        & \textit{30.1} & \textbf{28.8}        & 28.1        & \textit{27.5} & \textit{34.5} & 27.0        & 29.2        \\
\textbf{IB+M2O} & 244M & \textbf{24.8}        & \textbf{33.9}        & \textbf{37.2}        & \textbf{32.4}          & 28.5        & \textbf{28.5}        & \textbf{28.8}          & \textbf{35.7}          & \textbf{27.3}        & \textbf{29.5}        \\
\textbf{IALB+M2O} & 97M & 23.1 & 33.2 & 34.4 & 29.5 & 27.1 & 27.0 & 27.3 & 34.1 & 25.2 & 27.4\\ \hline
\multicolumn{12}{c}{\textbf{Distilled Large Models}}  \\\hline
\textbf{MB50+M2O}   &  611M   & {26.1} & {35.9} & {38.3} & \textit{32.9} & 29.6 & 29.3 & \textit{30.1} & \textit{37.1} & 28.5 & 31.7        \\
\textbf{IB+M2O}   & 244M     & 26.0 & {35.9} & 38.0 & {33.7} & {29.9} & {29.4} & {30.3} & {37.4} & {28.4} & {31.6}        \\\hline
\multicolumn{12}{c}{\textbf{Distilled Compact Models}}  \\\hline
\textbf{M2O}  & 78M      & 23.6 & 33.3 & 36.0 & 30.2 & 26.0 & 26.9 & 27.7 & 34.0 & 25.6 & 27.8        \\
\textbf{IAIB+M2O}   & 97M     & \textbf{24.9} & \textbf{34.4} & \textbf{36.6} & \textbf{31.9} & \textbf{27.7} & \textbf{28.1} & \textbf{28.6} & \textbf{35.5} & \textbf{26.5} & \textbf{29.0}        \\\hline
                            \multicolumn{12}{c}{\textbf{En-XX}}\\\cline{1-12}                                                                                                
\multicolumn{12}{c}{\textbf{Bilingual Models}}  \\\hline             
\textbf{Bi}       & 78M         & 4.5         & 17.9        & 21.7        & 12.1          & 3.9         & 10.0        & 9.2           & 17.9          & 7.2         & 2.1         \\
\textbf{MB50+Bi}    & 611M    & \textbf{8.6}         & 23.5        & \textbf{27.0}        & \textit{17.4} & \textbf{6.0}         & \textbf{15.8}        & \textit{11.6} & \textit{24.5} & \textbf{11.2}        & 3.3         \\
\textbf{IB+Bi}  & 244M & 8.2         & \textbf{23.6}        & 26.9        & \textbf{17.7}        & \textbf{6.0}         & \textbf{15.8}        & \textbf{11.8}        & \textbf{25.1}        & 10.8        & \textbf{3.6}         \\\hline
\multicolumn{12}{c}{\textbf{Multilingual Models}}  \\\hline
\textbf{O2M}        & 78M       & 7.4         & 22.5        & 25.9        & 16.2          & 5.6         & 14.7        & 11.4          & 21.9          & 10.0        & 2.7         \\

\textbf{MB50+O2M}    & 611M    & 8.9         & 22.8        & \textbf{27.5} & \textit{18.1} & 6.5         & 16.3        & \textit{12.0} & \textit{25.1} & \textbf{11.6}        & \textbf{3.7}         \\
\textbf{IB+O2M}& 244M & \textbf{9.1}         & \textbf{24.0}          & 27.3        & \textbf{18.5}          & \textbf{6.7}         & \textbf{16.7}        & \textbf{12.9}          & \textbf{26.4}          & \textbf{11.6}        & \textbf{3.7}  \\
\textbf{IALB+O2M} & 97M & 8.1 & 22.3 & 26.3 & 17.0 & 5.8 & 15.3 & 11.6 & 24.2 & 10.5 & 3.2 \\\hline
\multicolumn{12}{c}{\textbf{Distilled Large Models}} \\\hline
\textbf{MB50+O2M}    & 611M    & \textbf{9.4} & 24.5 & 27.5 & \textit{17.5} & 6.1 & 16.4 & \textit{12.8} & \textit{26.3} & 11.6 & 2.9 \\
\textbf{IB+O2M}  & 244M      &      9.3 & \textbf{25.0} & \textbf{28.2} & \textbf{19.2} & \textbf{6.7} & \textbf{17.0} & \textbf{13.2} & \textbf{26.5} & \textbf{11.8} & \textbf{3.7}\\ \hline
\multicolumn{12}{c}{\textbf{Distilled Compact Models}} \\\hline
\textbf{O2M}   & 78M     &    \textbf{8.9} & \textbf{24.1} & \textbf{27.5} & \textbf{18.2} & \textbf{6.3} & 16.0 & 12.5 & \textbf{25.6} & 11.0 & \textbf{3.2}     \\
\textbf{IAIB+O2M}  & 97M      &   \textbf{8.9} & 23.4 & 27.2 & 17.8 & \textbf{6.3} & \textbf{16.2} & \textbf{12.7} & 25.3 & \textbf{11.3} & 3.1      \\
\end{tabular}
\caption{Comparison of IndicBART with other models. Scores are reported on the WAT 2021 test set.}
\label{tab:bigpicture}
\end{table*}




\subsection{Model Training Settings}

We use a single GPU for bilingual and 8 GPUs for multilingual models, all of which are Transformers. Multilingual models are trained using the approach in \citet{johnson-etal-2017-googles} where corpora for various language pairs are first balanced according to their size, then concatenated after appending target language indicator tokens, and finally fed to the NMT model for training. Wherever possible and applicable, we tuned hyperparameters such as hidden sizes, dropout, label smoothing, warm-up, tokens per batch, per GPU, learning rate and weight decay. The ADAM optimizer was used. We train our models till convergence on the development set BLEU scores \cite{papineni}. We decode train/tests sets using beam search with a beam of size 4 and a length penalty of 0.8. We report the BLEU scores on the decoded results computed using sacreBLEU\footnote{BLEU+case.mixed+numrefs.1+smooth.exp+tok.13a +version.1.5.1} \cite{post-2018-call}. For additional details, refer to section~\ref{supp:trainsettings} in the appendix.

\subsection{Comparison of Pre-trained Models}
We first describe the main results of using IndicBART and its variants for machine translation and compare it with other relevant models. Table~\ref{tab:bigpicture} shows results for models trained on the PMI corpus and evaluated on the WAT21 test set.

\noindent \textbf{Language specific models are compact and competitive:}  Considering bilingual models, IndicBART outperforms models trained from scratch and gives competitive results when compared to mBART50. For Indic to English translation, mBART50 tends to be better, but this is not surprising because it is trained on far larger amounts of English data in addition to being  almost 3 times larger than IndicBART. For English to Indic translation, both models tend to give similar scores. In the case of multilingual models, IndicBART is, once again, vastly better than its counterpart trained from scratch and when compared to mBART50 the gap which existed in case of bilingual settings disappears and sometimes reverses in favor of IndicBART. In both cases, IndicBART outperforms mBART50 for Kannada, Punjabi and Oriya which the latter is not trained for. This shows that having a compact language group specific model can be competitive with if not better than a general purpose model trained on a larger number of  languages while only having one-third the number of parameters as the latter.

\noindent \textbf{Extreme compression has its downside:} Comparing the performance of IndicBART and mBART50 against IndicALBART in multilingual settings, it seems that a 60\% and 84\% reduction of parameters, respectively, has a negative impact on the translation quality, which results in drops of up to 3 BLEU. However, this may be considered as a reasonable tradeoff given the high levels of compression achieved. Especially given that IndicALBART is 84\% smaller than mBART50, means that large capacity GPUs (which not everyone has easy access to) may not be needed. Furthermore, the drops in quality can be addressed via distillation.

\noindent \textbf{Distillation successfully transfers performance from large to smaller models:} We see that fine-tuning the pre-trained IndicALBART on distilled data from IndicBART can match the performance of the IndicBART model. Fine-tuning pre-trained IndicALBART performs better than training a randomly initialized model on the same distilled data in the XX-En direction. On the other hand, both the approaches are competitive in the En-XX direction.

\noindent \textbf{Self-training on distilled data is beneficial:} When IndicBART and MB50 are fine-tuned on distillation data generated from a previously fine-tuned model, we see significant improvements in the XX-En direction, and modest improvements in the En-XX directions. These observations are mostly in line with \citet{dabre-fujita-2020-combining}.


In summary, compact language group specific pre-trained models are competitive with large universal language models. This can result in reasonable gains in fine-tuning multilingual models (3.3-3.5 hours for IndicBART variants vs 4.7-5 hours for mBART50) and  significantly reduce the memory footprint (97-244M vs 611M) for deployment.



\begin{table}[t]
\centering
\resizebox{\columnwidth}{!}{%
\begin{tabular}{l|ccccc}
\multirow{2}{*}{\textbf{Model}}    & \textbf{bn} & \textbf{hi} & \textbf{ml} & \textbf{or} & \textbf{ta}                                                                                             \\\cline{2-6}
                           &  \multicolumn{5}{c}{\textbf{XX-En}} \\\hline
\textbf{IB+M2O} & \textbf{24.8}               & \textbf{37.2}                & \textbf{28.5}                & \textbf{28.8}               & \textbf{27.3}         \\
\textbf{SSIB+M2O} & 24.1            & 35.5               & 27.9            & 28.1                & 26.9                \\\cline{1-6}
                           & \multicolumn{5}{c}{\textbf{En-XX}}                                                                                             \\\cline{1-6}
                          
\textbf{IB+O2M} & \textbf{9.1}         &  \textbf{27.3}                & \textbf{6.7}                 & \textbf{16.9}                & \textbf{11.6}                 \\
\textbf{SSIB+O2M} & 9.3                 & \textbf{27.3}                & 6.2                 & 16.6                & 11.4                 \\
\end{tabular}
}
\caption{Ablation studies on the impact of multilingualism and script unification on downstream performance of IndicBART. Scores are on the WAT 2021 test set.}
\label{tab:ablation}
\end{table}

\subsection{Ablation Studies}
We now perform ablation experiments to study the (a.)~impact of script unification on translation, (b.)~impact of corpora sizes and domains on translation, (c.)~translation quality for languages unseen during fine-tuning, and (d.)~translation quality on languages unseen during pre-training. Although we train models on all languages, we only report on a subset due to lack of space. Please see Sections~\ref{supp:scriptuni}, ~\ref{supp:corpsizedomain} in the appendix for more detailed results.

\subsubsection{Impact Of Script Unification}
Table~\ref{tab:ablation} contains the ablation tests, giving the results for the impact of script unification with multilingual fine-tuning. Comparing scores of models fine-tuned on unified script IndicBART (IB+M2O/O2M) against separate script IndicBART (SSIB+M2O/O2M) it is clear that overall, the former is better than the latter which could indicate that script unification enables languages to better benefit from each other. The case of Kannada, Punjabi and Oriya, further, illustrates the utility of script unification. The results for these languages are italicized in the rows labelled MB50+Bi and MB50+O2M/M2O in Table~\ref{tab:bigpicture}. mBART50 was not pre-trained on these languages, so we converted the training data in these languages in the Devanagari script\footnote{None of the pre-training languages use the same script as kn, pa, or.}. With this trick, we still managed to get large performance improvements over the baselines trained from scratch, and these improvements are often close to those exhibited by using IndicBART. This shows that we may not need to pre-train on all languages. However, explicitly training on the languages of interest should lead to better translation quality \cite{DBLP:journals/corr/abs-2008-00401}.

\begin{table}[t]
\centering
\resizebox{\columnwidth}{!}{%
\begin{tabular}{l|cccccccccc}
 \multirow{1}{*}{\textbf{Model}} & \textbf{bn} & \textbf{hi} & \textbf{ml} & \textbf{or} & \textbf{ta}                                                            \\\cline{1-6}
                             & \multicolumn{5}{c}{\textbf{Test Set: WAT 2021}}                                                    \\\cline{2-6}
            
\textbf{IB+PMI} & 24.8               & 37.2            & 28.5               & 28.8            & 27.3               \\
\textbf{IB+PMI+PIB}                 & \textbf{28.9}                                                  & \textbf{41.7}  & \textbf{33.2}  & \textbf{33.2}  & \textbf{32.0} \\
\textbf{Samanantar}             & 27.9    & 41.8    & 32.7    & 32.9    & 31.2   \\
\textbf{IB+Samanantar}             &27.1    & 41.0    & 31.6    & 32.3    & 30.1  \\\hline\hline
& \multicolumn{5}{c}{\textbf{Test Set: FLORES}}                                           \\\cline{2-6}
\textbf{IB+PMI}   & 10.4    & 14.8    & 8.1    & 11.2    & 10.5  \\
\textbf{IB+PMI+PIB}   & 13.0    & 22.0    & 12.7    & 15.1    & 13.8  \\
\textbf{Samanantar}   & \textbf{30.7}    & \textbf{36.0}    & \textbf{30.4}    & \textbf{28.6}    & \textbf{27.7}  \\
\textbf{IB+Samanantar}   & 30.1    & 35.3    & 29.1    & 28.5    & 26.6  \\

\end{tabular}
}
\caption{Ablation study of the impact of using different fine-tuning corpora sizes (PMI+PIB) and their comparison against a model trained from scratch as well as fine-tuned on a general domain corpus (Samanantar). We evaluate Indic to English translation on the WAT 2021 as well as the FLORES test sets.}
\label{tab:corporasize}
\end{table}


\subsubsection{Impact Of Corpora Size and Domain}\label{sec:sizedomain}
Table~\ref{tab:corporasize} shows the impact of corpora sizes as well as training data domain on some Indic to English pairs (complete results in  Appendix~\ref{supp:corpsizedomain}). All models are multilingual (M2O), have the same size and are trained on unified script data. In order to clearly assess the impact of domains, we evaluate on the WAT 2021 as well as the FLORES test sets. 
Regardless of the test sets or testing domains, comparing rows IB+PMI and IB+PMI+PIB, it is clear that increasing the amount of fine-tuning data has a positive impact on the final translation quality. However, PMI+PIB data is in-domain for the WAT 2021 test set but out-of-domain for the FLORES test set, and the performance on the latter test set still improves.
Furthermore, comparing rows IB+PMI+PIB and Samanantar, we can see widely different results depending on the test set. For the WAT 2021 test set, fine-tuning on the PMI+PIB dataset is comparable to training on Samanantar from scratch, indicating that for domain specific models, having a small in-domain fine-tuning data is sufficient. On the other hand, on the more general domain FLORES test sets training on the more diverse Samanantar data is clearly better. Finally, the scores in the row IB+Samanantar show that pre-training has minimal impact when the parallel corpora are large, an observation in line with \citet{liu-etal-2020-multilingual-denoising}. 


\begin{table}[t]
\centering
\resizebox{\columnwidth}{!}{%
\begin{tabular}{l|cc|cc}
\multirow{2}{*}{\begin{tabular}[c]{@{}l@{}}\textbf{Setting}\end{tabular}} &\multicolumn{2}{c}{\textbf{M2O}} & \multicolumn{2}{c}{\textbf{O2M}}\\
 & \multirow{1}{*}{\textbf{kn-en}} & \multirow{1}{*}{\textbf{pa-en}} & \multirow{1}{*}{\textbf{en-kn}} & \multirow{1}{*}{\textbf{en-pa}} \\\hline
                               
\textbf{IB+Full}               & 32.4                          & 35.7                      & 18.5 & 26.4   \\\hline
\textbf{IB+Zero}           & 27.5                         & 31.5 & 6.1 & 10.4             \\
\textbf{SSIB+Zero}           & 24.0                         & 28.2 & 3.9 & 7.4              \\
\end{tabular}
}
\caption{Evaluation of Kannada and Punjabi to/from English translation, which aren't seen when fine-tuning.}
\label{tab:knpa}
\end{table}

\subsubsection{Unseen Languages During Fine-Tuning}
We evaluate Kannada and Punjabi to/from English translation where the IndicBART model, with and without script unification, is fine-tuned on the multilingual PMI data where the training data for these languages is missing (denoted by ``Zero''). We compare against a setting where the training data is used (denoted by ``Full''). Table~\ref{tab:knpa} shows what happens when languages are seen during pre-training but not during fine-tuning. There are two critical observations: First, despite not having seen any training data for the given language pairs, we still obtain a reasonable translation for translation into English. However, the quality of translation from English is poor due to the decoder not having seen those specific Indic languages during fine-tuning. Incorporating a monolingual de-noising objective for unseen target languages during fine-tuning could alleviate this problem. Second, script unification has a large impact on the final performance, often improving performance by up to 3.5 BLEU over a separate script model.

\subsubsection{Unseen Languages During Pre-Training}
We study Nepalese (ne) and Sinhala (si) to English translation using the parallel training data from \citet{guzman2019flores} (also used in \citet{liu-etal-2020-multilingual-denoising}) for bilingual fine-tuning, and evaluate on the FLORES devtest set\footnote{\url{https://github.com/facebookresearch/flores}}. Note that for Sinhala we have to resort to script mapping into Devanagari.
Table~\ref{tab:nesi} shows what happens when we perform fine-tuning for languages that IndicBART is not trained on. The baselines, trained using the unified script IndicBART vocabulary, will seem weaker than what is reported in previous work, but it should be noted that the vocabulary was not actually trained for Nepali and Sinhala. Regardless, fine-tuning leads to substantial improvements in translation quality, which indicates the utility of IndicBART even for unseen languages. Comparing against \citet{liu-etal-2020-multilingual-denoising} who use the same fine-tuning data as us but their mBART model is pre-trained on both languages, we can see that our models are not too far behind. 

\begin{table}[t]
\centering

\begin{tabular}{l|cc}
\multirow{1}{*}{\textbf{Model}} & \multirow{1}{*}{\textbf{ne-en}} & \multirow{1}{*}{\textbf{si-en}} \\\hline
                               
\textbf{Bi (Scratch)}               & 5.2                          & 4.3                          \\
\textbf{IB+Bi}           & \textbf{10.5}                         & \textbf{8.5}                         \\\hline
\cite{liu-etal-2020-multilingual-denoising}           & 14.5                         & 13.7                         
\end{tabular}
\caption{Evaluation of Nepali and Sinhala to English translation where IndicBART hasn't seen Nepali and Sinhala during pre-training.}
\label{tab:nesi}
\end{table}

\section{Experiments: Extreme Summarization} 
We compare the performance of fine-tuning IndicBART, its variants and mBART50 on the challenging \textit{extreme summarization} task \cite{narayan-etal-2018-dont} for Indic languages. The small datasets, enable a good study of the utility of pre-training.

\subsection{Models Trained}
We fine-tune and compare the mBART50 (MB), IndicBART (IB), IndicALBART (IALB) and the separate script IndicBART model (SSIB) models. Punjabi is not present in mBART50 and has its script mapped to Devanagari before fine-tuning (italicized results).  

\subsection{Datasets and Preprocessing}
We used the multilingual XL-Sum dataset \cite{hasan-etal-2021-xl} for our experiments. The Indic languages we focus on for evaluating our IndicBART models are: Bengali, Gujarati, Hindi, Marathi, Punjabi, Tamil and Telugu. 
We use the updated splits of \citet{hasan-etal-2021-xl}, the statistics of which are given in their GitHub page\footnote{\url{https://github.com/csebuetnlp/xl-sum/}}. 
Since the splits are not n-way parallel, we do not conduct multilingual fine-tuning due to potential content overlaps between splits across languages. Like we did in NMT, we map all scripts to Devanagari as applicable for fine-tuning (only Punjabi for mBART50, all languages for IndicBART and IndicALBART and none for separate script IndicBART). Statistics are given in Table~\ref{tab:xsumdata} in the appendix.

\begin{table}[t]
    \centering
    \begin{tabular}{l|cccc}
         
        \multirow{2}{*}{\textbf{Lang}}  & \multirow{2}{*}{\textbf{MB50}}  & \multirow{2}{*}{\textbf{IB}} & \multirow{2}{*}{\textbf{SSIB}}   & \multirow{2}{*}{\textbf{IALB}}\\\\
         \hline
          bn & \textbf{21.87}   & 21.46  & 20.52 & 19.86 \\
          gu &  \textbf{18.28}   &  18.20 & 16.38 & 16.81 \\
          hi &   \textbf{31.71}  &  30.94 & 30.33 & 30.04 \\
          mr & 18.33   &  \textbf{19.00} & 18.66 & 18.44 \\
        
          pa &   \textit{22.14}  &    24.82 & \textbf{25.08} & 23.29\\
       
          ta & 19.50   &  \textbf{20.40} & 20.23 & 17.41\\
          te & 13.34  & \textbf{14.38} & 13.34 & 13.55\\

    \end{tabular}
    \caption{Rouge-L scores for summarization on XL-Sum.}  
    \label{tab:xsum-results}
\end{table}

\subsection{Model Training Settings}
Similar to NMT, we use YANMTT for fine-tuning. We use maximum document-summary lengths of 512-64 tokens, which loosely follows previous work \cite{lewis-etal-2020-bart}. Most of the optimal hyperparameters were the same as for NMT. We train our models till convergence on the development set Rouge-L F1 scores (RL) \cite{lin-2004-rouge}. For decoding test sets, we use beam size of 5, length penalty of 1.2 and a decoding n-gram repetition limit of 4\footnote{This means that 4-grams won't be repeated in the output.}. We report RL scores on the decoded results computed using multilingual Rouge scoring toolkit\footnote{\url{https://github.com/csebuetnlp/xl-sum/tree/master/multilingual_rouge_scoring}}. Refer to section~\ref{supp:summtrain} in the appendix for details.

\subsection{Results}
Table \ref{tab:xsum-results} contains the results for the summarization experiments. IndicBART (IB) and mBART50 are competitive with each other where the former performs slightly better for  Marathi, Punjabi, Tamil and Telugu. Once again, separate script IndicBART (SSIB) fared poorer than IndicBART except for Punjabi, indicating the importance of script unification. Similar to NMT, fine-tuning IndicALBART gives poorer results, often lagging 1-3 RL points behind IndicBART which we consider to be a reasonable tradeoff given the reduced parameter sizes.  We expect that distillation may help improve performance, like it does for NMT. Overall, the major conclusions are in line with the those observed for the low-resource NMT task.


\section{Conclusion and Future Work}

We presented IndicBART, a multilingual, pre-trained sequence-to-sequence model to support development of NLG applications for Indian languages. IndicBART supports 11 Indian languages and English, and utilizes the orthographic similarity of Indic scripts to enable better cross-lingual transfer. IndicBART presents a case-study for language group-specific pre-trained S2S models. Our experiments on fine-tuning IndicBART for NMT and summarization showed that the model is competitive with large models such as mBART50. We further  compressed IndicBART while maintaining downstream task performance via parameter sharing (IndicALBART) combined with  multilingual distillation. We showed that script unification has a strong positive impact on translation and summarization. We also showed that IndicBART, thanks to its script independent nature, can be readily used for enabling translation for languages such as Sinhala and Nepali which IndicBART has not been explicitly pre-trained for. Furthermore, we showed that fine-tuning IndicBART on one set of languages enables translation for another unseen set of languages, which shows that pre-trained models enable translation without parallel corpora.

In the future, we plan to support more Indic languages in IndicBART; starting with all the 22\footnote{\url{https://www.mha.gov.in/sites/default/files/EighthSchedule_19052017.pdf}} languages listed in the $8^{th}$ schedule of the Indian constitution. Increased language coverage and models with lower compute demands can democratize access to NLP technologies.
We also plan to focus on training models on longer text chunks (documents) and larger text corpora, incorporating advances in multilingual pre-training, cross-lingual transfer and cross-lingual tasks for Indian languages.

\bibliography{anthology,indicnlg}
\bibliographystyle{acl_natbib}

\appendix

\section{Corpora statistics}
Table~\ref{tab:databi} gives the statistics for the monolingual corpora, Indiccorp (IC), and parallel corpora, PMI, PIB and Samanantar (Sam) used in this paper. Indiccorp is used for pre-training IndicBART and the parallel corpora are used for fine-tuning or for training models from scratch. PMI and PIB have similar domains. PMI is used to simulate a realistic low-resource domain specific setting, and PIB is used to simulate a middle-resource domain specific setting. Samanantar is used to simulate a high resource general domain setting.

\begin{table}[t]
\centering
\resizebox{\columnwidth}{!}{%
\begin{tabular}{l|r|rr|r|r}
\multirow{3}{*}{\textbf{Lang}}& \multirow{2}{*}{\textbf{Mono}} & \multicolumn{4}{c}{\textbf{Parallel (XX-En)}} \\  \cline{3-6}
& & \multicolumn{3}{c|}{\textbf{LR}} & \multicolumn{1}{c}{\textbf{HR}}\\ \cline{2-6}
 & \textbf{IC} & \textbf{PMI} & \textbf{PIB} & \textbf{Total} & \textbf{Sam}\\ \hline
as       &  1.4M    & -       & -  & - &  -\\
bn       &    39.9M  & 23.3K       & 91.9K  & 115.2K &  8.4M\\
en       &    54.3  & -       & -  & - &  -\\
gu          &  41.1M  & 41.5K        & 58.2K  & 99.8K &  3.0M\\
hi      & 63.1M       & 50.3K        & 266.5K  & 316.9K &  8.4M\\
kn      & 53.3M       & 28.9K        & -   & 28.9K &4.0M\\
ml      & 50.2M       & 26.9K        & 43.1K   & 70.0K  & 5.8M\\
mr      & 34.0M       & 28.9K        & 114.2K  & 143.1K  & 3.2M\\
or      & 7.0M       & 31.9K        & 94.4K   & 126.4K & 990.4K\\
pa      & 29.2M       & 28.2K        & 101,092  & 129.3K  & 2.4M\\
ta      & 31.5       & 32.6K        & 115.9K  & 148.6K &  5.1M\\
te      & 47.9M       & 33.3K & 44.7K & 78.1K &4.7M\\
\hline
Total         & 450M   &  326.3K      & 930.3K & 1.2M & 46.2M\\

\hline
\end{tabular}}
\caption{Statistics of monolingual and parallel corpora (\#sentences) for pre-training IndicBART and fine-tuning it, respectively.}
\label{tab:databi}

\end{table}

\section{NMT Model Training Settings}
\label{supp:trainsettings}
We use a single GPU for bilingual and 8 GPUs for multilingual models, all of which are Transformers. Multilingual models are trained using the approach in \citet{johnson-etal-2017-googles}. Due to the large number of models we train, we did not perform exhaustive hyperparameter tuning. We mainly focused on tuning the learning rates, batch sizes and warm-ups. We found that high dropouts were surprisingly ineffective, especially for multilingual settings, regardless of training from scratch or fine-tuning. Nevertheless, for fine-tuning IndicBART and its variants, we determined the following optimal hyperparameters: dropouts of 0.1, label smoothing of 0.1, warm-up of 16,000 steps, 2048 tokens per batch per GPU, learning rate of 0.001 and weight decay of 0.00001 with the ADAM optimizer for training. For mBART50, we used warm-up of 2,500 steps, 512 tokens per batch per GPU, and a learning rate of 0.00003.\footnote{A small learning rate is needed since we can train on very small batches given the large model size.} For bilingual and multilingual models trained from scratch on the small PMI and PIB data, we use smaller models with hidden and filter sizes of 512 and 2048, respectively, while keeping all other hyperparameters the same as for IndicBART which we found to be highly effective. As Samanantar data is much larger, we keep its size the same as IndicBART. Except for separate script IndicBART and mBART50, all models use the same vocabulary as IndicBART for consistency.

We train our models till convergence on the development set BLEU scores \cite{papineni} which are computed via greedy decoding every 1,000 batches. For multilingual models, we use the global development set BLEU score, an average of BLEU scores for each language pair. During decoding the test sets, we use beam search with a beam of size 4 and a length penalty of 0.8. We report the BLEU scores on the decoded results computed using sacreBLEU\footnote{BLEU+case.mixed+numrefs.1+smooth.exp+tok.13a +version.1.5.1} \cite{post-2018-call}.

\begin{table*}[ht]
\centering
\begin{tabular}{l|cccccccccc}
\multirow{2}{*}{\textbf{Model}}    & \textbf{bn} & \textbf{gu} & \textbf{hi} & \textbf{kn} & \textbf{ml} & \textbf{mr} & \textbf{or} & \textbf{pa} & \textbf{ta} & \textbf{te}                                                                                             \\\cline{2-11}
                          &  \multicolumn{10}{c}{\textbf{XX-En}} \\\hline
\textbf{IB+M2O} & \textbf{24.8}        & \textbf{33.9}        & \textbf{37.2}        & \textbf{32.4}        & \textbf{28.5}        & \textbf{28.5}        & \textbf{28.8}        & \textbf{35.7}        & \textbf{27.3}        & 29.5        \\
\textbf{IB$^{no SM}$+M2O} & 24.1        & 33.8        & 35.5        & 31.2        & 27.9        & 28.0        & 28.1        & \textbf{35.7}        & 26.9        & 28.4        \\
\textbf{IB+Bi}   & 23.6        & 35.5        & 36.8        & 31.6        & 27.9        & 26.8        & 28.3        & 36.3        & 27.0        & \textbf{29.9}        \\
\textbf{IB$^{no SM}$+Bi}   & 22.3        & 34.9        & 36.6        & 30.8        & 27.5        & 26.7        & 28.0        & 36.0        & 26.3        & 29.7        \\\cline{1-11}
                          & \multicolumn{10}{c}{\textbf{En-XX}}                                                                                             \\\cline{1-11}
                          
\textbf{IB+O2M} & \textbf{9.1}         & \textbf{24.0}        & \textbf{27.3}        & 18.5        & \textbf{6.7}         & \textbf{16.7}        & 12.9        & \textbf{26.4}        & \textbf{11.6}        & \textbf{3.7}         \\
\textbf{IB$^{no SM}$+O2M} & 9.3         & \textbf{24.0}        & \textbf{27.3}        & 17.9        & 6.2         & 16.4        & \textbf{16.6}        & 23.4        & 11.4        & 3.0         \\
\textbf{IB+Bi}   & 8.2         & 23.6        & 26.9        & 17.7        & 6.0         & 15.8        & 11.8        & 25.1        & 10.8        & 3.6         \\
\textbf{IB$^{no SM}$+Bi}   & 8.2         & 22.9        & 26.6        & 17.3        & 5.8         & 14.6        & 14.8        & 22.9        & 10.5        & 3.6        
\end{tabular}
\caption{Ablation studies to study the impact of multilingualism and script unification on downstream performance of IndicBART. Scores are reported on the WAT 2021 test set.}
\label{tab:suppablation}
\end{table*}

\section{NMT Results: Impact of Script Unification}
\label{supp:scriptuni}
Table~\ref{tab:suppablation} contains the results of ablation studies on the impact of script unification in bilingual and multilingual settings. Regardless of bilingual or multilingual fine-tuning, it is clear that script unification tends to give better results on average as compared to using separate scripts to represent all languages.

\begin{table*}[!ht]
\centering
\begin{tabular}{l|cccccccccc}
 & \multicolumn{10}{c}{\textbf{Test Set: WAT 2021}}                                           \\\cline{2-11}
\multirow{2}{*}{\textbf{Model}}     & \textbf{bn} & \textbf{gu} & \textbf{hi} & \textbf{kn} & \textbf{ml} & \textbf{mr} & \textbf{or} & \textbf{pa} & \textbf{ta} & \textbf{te}                                                                                             \\\cline{2-11}
                          &  \multicolumn{10}{c}{\textbf{XX-En}} \\\hline
\textbf{IB+PMI} & 24.8        & 33.9        & 37.2        & 32.4        & 28.5        & 28.5        & 28.8        & 35.7        & 27.3        & 29.5        \\
\textbf{IB+PMI+PIB}                 & \textbf{28.9}                         & 38.8                         & \textbf{41.7} & 34.6 & \textbf{33.2} & \textbf{32.5} & \textbf{33.2} & 41.3 & \textbf{32.0} & 33.0 \\
\textbf{Samanantar}             & 27.9  & \textbf{39.0}  & 41.8  & \textbf{34.8}  & 32.7  & 32.0  & 32.9  & \textbf{41.4}  & 31.2  & \textbf{34.4} \\
\textbf{IB+Samanantar}             &27.1  & 38.0  & 41.0  & 34.1  & 31.6  & 31.1  & 32.3  & 40.1  & 30.1  & 32.4\\\hline

                          & \multicolumn{10}{c}{\textbf{En-XX}}                                                                                             \\\hline

\textbf{IB+PMI} & 9.1         & 24.0        & 27.3        & 18.5        & 6.7         & 16.7        & 12.9        & 26.4        & 11.6        & 3.7         \\
\textbf{IB+PMI+PIB}                 & \textbf{11.1}                         & \textbf{25.5}                         & \textbf{33.0} & \textbf{18.9} & \textbf{7.2} & \textbf{19.1} & \textbf{14.3} & \textbf{27.1} & \textbf{13.6} & 3.6 \\
\textbf{Samanantar}             & 9.7  & 24.7  & \textbf{33.0}  & 17.5  & 7.0  & 18.4  & 13.3  & 25.5  & 12.7  & \textbf{5.8}\\
\textbf{IB+Samanantar}             & 9.4  & 24.2  & \textbf{33.0}  & 17.2  & 6.5  & 17.7  & 13.5  & 25.6  & 11.8  & 5.6\\\hline\hline
& \multicolumn{10}{c}{\textbf{Test Set: FLORES}}                                           \\\cline{2-11}
\multirow{2}{*}{\textbf{Model}}     & \textbf{bn} & \textbf{gu} & \textbf{hi} & \textbf{kn} & \textbf{ml} & \textbf{mr} & \textbf{or} & \textbf{pa} & \textbf{ta} & \textbf{te}                                                                                             \\\cline{2-11}
                          &  \multicolumn{10}{c}{\textbf{XX-En}} \\\hline
                          
\textbf{IB+PMI}   & 10.4  & 13.2  & 14.8  & 11.8  & 8.1  & 10.1  & 11.2  & 12.9  & 10.5  & 10.5\\
\textbf{IB+PMI+PIB}   & 13.0  & 18.4  & 22.0  & 13.1  & 12.7  & 16.1  & 15.1  & 18.5  & 13.8  & 16.2\\
\textbf{Samanantar}   & \textbf{30.7}  & \textbf{33.6}  & \textbf{36.0}  & \textbf{27.4}  & \textbf{30.4}  & \textbf{30.0}  & \textbf{28.6}  & \textbf{34.2}  & \textbf{27.7}  & \textbf{32.7}\\
\textbf{IB+Samanantar}   & 30.1  & 32.6  & 35.3  & 27.2  & 29.1  & 29.6  & 28.5  & 33.0  & 26.6  & 32.1\\\hline

                          & \multicolumn{10}{c}{\textbf{En-XX}}                                                                                             \\\hline

\textbf{IB+PMI}   & 3.5  & 9.5  & 14.7  & 5.6  & 2.1  & 6.0  & 5.3  & 10.6  & 5.0  & 3.1\\
\textbf{IB+PMI+PIB}   & 5.4  & 13.5  & 22.8  & 7.5  & 2.8  & 9.1  & 6.4  & 15.5  & 6.9  & 3.5\\
\textbf{Samanantar}   & \textbf{17.3}  & \textbf{22.6}  & \textbf{31.3}  & \textbf{16.7}  & \textbf{14.2}  & \textbf{14.7}  & \textbf{10.1}  & \textbf{21.9}  & \textbf{14.9}  & \textbf{20.4}\\
\textbf{IB+Samanantar}   & 17.1  & 21.5  & 31.2  & 16.2  & 13.0  & 14.2  & 10.2  & 21.5  & 13.7  & 19.5\\\\
\end{tabular}
\caption{Ablation study of the impact of using different sizes of fine-tuning corpora (PMI and its combination with PIB) and their comparison against a model trained from scratch as well as fine-tuned on a general domain corpus (Samanantar). We evaluate on the WAT 2021 as well as the FLORES test sets.}
\label{tab:suppcorporasize}
\end{table*}

\section{NMT Results: Effect of Corpora Size and Domain}
\label{supp:corpsizedomain}

Table~\ref{tab:suppcorporasize} contains the results showing the impact of varying corpora sizes and domain on translation quality. In the main paper, we could not show results for all languages and directions, due to lack of space. There are three key points to note: (a.) fine-tuning using small in-domain corpora (PMI) gives competitive results compared to using a large general domain corpus. (b.) Additional corpora from a related domain (PMI) leads to substantial improvements in translation quality for in- as well as out-of-domain performance, indicating that fine-tuning a pre-trained model on a corpus belonging to a different domain (PMI/PIB) is a viable option in case training corpus for the target domain (FLORES) is unavailable. Furthermore, going from low-resource to middle resource settings does not diminish the contribution of pre-trained models. (c.) General domain corpora inevitably lead to the best performance, but since training large models on large general domain corpora is more time-consuming, fine-tuning is a more attractive option since pre-training needs to be done only once.

\begin{table}[!ht]
    \centering
    \begin{tabular}{c|ccc}
\textbf{Language} & \textbf{Train} & \textbf{Dev} & \textbf{Test} \\\hline
bn & 8,102 & 1,012 & 1,012 \\
gu & 9,119 & 1,139 & 1,139 \\
hi & 70,778 & 8,847 & 8,847 \\
mr & 10,903 & 1,362 & 1,362 \\
pa & 8,215 & 1,026 & 1,026 \\
ta & 16,222 & 2,027 & 2,027 \\
te & 10,421 & 1,302 & 1,302 \\\hline

    \end{tabular}
    \caption{Statistics of the Indic portion of the multilingual XL-Sum dataset \cite{hasan-etal-2021-xl} that we used for training our summarization models.}
    \label{tab:xsumdata}
\end{table}

\section{Corpora statistics for summarization experiments}
Table~\ref{tab:xsumdata} contains statistics of the Indic section of the XL-sum dataset, which we use for summarization experiments. We preprocess languages by mapping their scripts to Devanagari as applicable (all languages for IndicBART and IndicALBART; none for separate script IndicBART; only Punjabi for mBART50).

\section{Summarization Model Training Settings }
\label{supp:summtrain}
Similar to NMT, we use YANMTT for fine-tuning. We use maximum document-summary lengths of 512-64 tokens, which loosely follows previous work \cite{lewis-etal-2020-bart}. Unlike NMT, we do not train models from scratch, as they would not work given the small data sizes and difficulty of summarization. For IndicBART and its variants, we determined the following optimal hyperparameters: batch sizes of 4,096 tokens, dropouts of 0.1, label smoothing of 0.1, learning rate warmup steps of 4,000, learning rate of 0.001 and weight decay of 0.00001 with the ADAM optimizer. For mBART50 we use sentence level batching with 2 document-summary pairs per batch and learning rate of 0.00001 which we found to be optimal. We train our models till convergence on the development set Rouge scores (Rouge-L F1) \cite{lin-2004-rouge} for all languages, which are computed via greedy decoding every 1,000 batches. Similar to NMT, we save the best performing checkpoints for each language. During decoding the test sets, we use beam search with a beam of size 5, length penalty of 1.2 and a decoding n-gram repetition limit of 4-grams\footnote{This means that 4-grams won't be repeated in the output.}. We report Rouge scores on the decoded results computed using multilingual Rouge scoring toolkit\footnote{\url{https://github.com/csebuetnlp/xl-sum/tree/master/multilingual_rouge_scoring}}.

\end{document}